\documentclass[%
 reprint,
superscriptaddress,
 amsmath,amssymb,
 aps,
 prl,
]{revtex4-2}

\usepackage{graphicx}
\usepackage{dcolumn}
\usepackage{bm}
\usepackage{mathtools}

\graphicspath{{figures/}}

\begin{document}


\title{Discovering Sparse Interpretable Dynamics from Partial Observations}

\author{Peter Y. Lu}
\email{lup@mit.edu}
\affiliation{%
 Department of Physics, Massachusetts Institute of Technology, Cambridge, MA 02139, USA
}%
\author{Joan Ari\~{n}o}%
\affiliation{%
 Department of Physics, Massachusetts Institute of Technology, Cambridge, MA 02139, USA
}%
\affiliation{
 Department of Physics, Universitat Polit\`{e}cnica de Catalunya, Barcelona, Spain
}
\author{Marin Solja\v{c}i\'{c}}%
\affiliation{%
 Department of Physics, Massachusetts Institute of Technology, Cambridge, MA 02139, USA
}%

\date{\today}

\begin{abstract}
Identifying the governing equations of a nonlinear dynamical system is key to both understanding the physical features of the system and constructing an accurate model of the dynamics that generalizes well beyond the available data. We propose a machine learning framework for discovering these governing equations using only partial observations, combining an encoder for state reconstruction with a sparse symbolic model. Our tests show that this method can successfully reconstruct the full system state and identify the underlying dynamics for a variety of ODE and PDE systems.
\end{abstract}

\maketitle


\emph{Introduction.}---Analyzing data from a nonlinear dynamical system to understand its qualitative behavior and accurately predict future states is a ubiquitous problem in science and engineering. In many instances, this problem is further compounded by a lack of available data and only partial observations of the system state, e.g.\ forecasting fluid flow driven by unknown sources or predicting optical signal propagation without phase measurements. This means that, in addition to identifying and modeling the underlying dynamics, we must also reconstruct the hidden or unobserved variables of the system state. While traditional approaches to system identification have had significant success with linear systems, nonlinear system identification and state reconstruction is a much more difficult and open problem \cite{Ljung1999}. Moreover, modeling nonlinear dynamics in a way that provides interpretability and physical insight is also a major challenge.

Modern machine learning approaches have made significant strides in black box predictive performance on many tasks \cite{Goodfellow-et-al-2016}, such as data-driven prediction of nonlinear dynamics \cite{RAISSI2019686,BERG2019239,JMLR:v19:18-046} including methods that only use partial observations \cite{9053035,doi:10.1063/5.0019309,NEURIPS2018_69386f6b,2004.06243}. However, because deep learning models often fail to take into account known physics, they require vast quantities of data to train and tend to generalize poorly outside of their training distribution. Standard deep learning models also lack the interpretability necessary for developing a detailed physical understanding of the system, although recent unsupervised learning approaches can help mitigate this problem \cite{PhysRevX.10.031056}. Introducing physical priors and building physics-informed inductive biases, such as symmetries, into neural network architectures can significantly improve the performance of deep learning models and provide a greater degree of interpretability \cite{PhysRevX.10.031056,2001.04385,yin2021augmenting,2104.13478}.

Recent data-driven nonlinear system identification methods based on Koopman operator theory offer a compelling alternative to deep learning approaches as well as a theoretical framework for incorporating neural networks into system identification methods \cite{Mauroy2020,9147729,NIPS2017_3a835d32,2102.12086}. However, these approaches still encounter barriers when dealing with certain types of nonlinear dynamics, such as chaos, which lead to a problematic continuous spectrum for the Koopman operator that cannot be modeled by a finite-dimensional linear system, although some progress has been made in addressing these limitations \cite{2102.12086,Brunton2017,Lusch2018}.

In this work, we choose to directly learn the symbolic governing equations of motion, which are often sparse and provide a highly interpretable representation of the dynamical system that also generalizes well. By fitting a symbolic model, we can capture the exact dynamics of the many physical systems in nature governed by symbolic equations. Previous work has shown that, by imposing a sparsity prior on the governing equations, it is possible to obtain interpretable and parsimonious models of nonlinear dynamics \cite{Brunton3932,doi:10.1098/rspa.2020.0279,2005.03448}. This sparsity prior, in combination with an autoencoder architecture, can also aid in extracting interpretable state variables from high dimensional data \cite{Champion22445}.

\begin{figure*}
\includegraphics{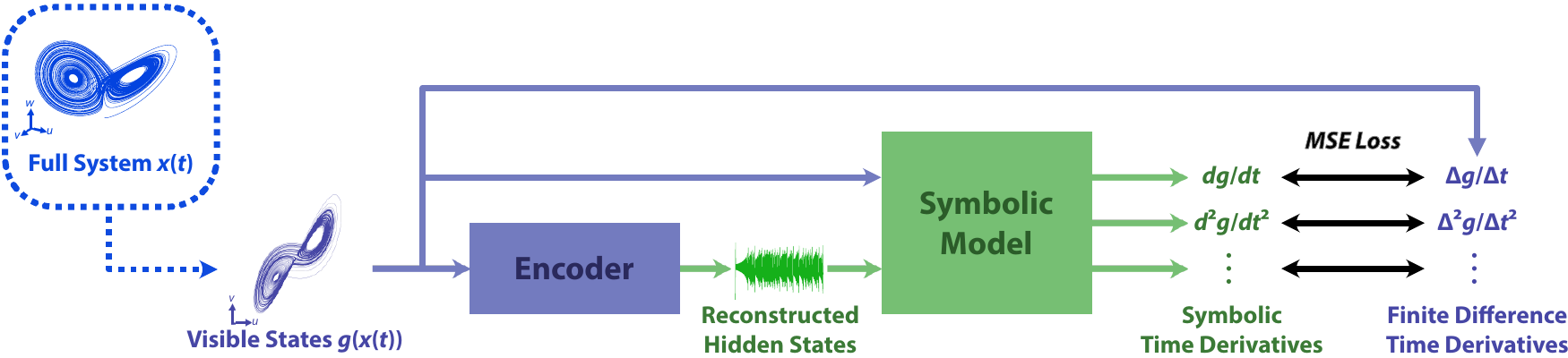}
\caption{\label{fig:arch}A machine learning framework for simultaneous system identification and state reconstruction. With only a visible portion of the full state available $\mathbf{x}_v = \mathbf{g}(\mathbf{x})$, an encoder is first used to reconstruct the hidden states. The fully reconstructed state $\hat{\mathbf{x}}$, including the visible and hidden states, is then passed into a symbolic model of the governing equations. Using automatic differentiation, multiple symbolic time derivatives $d^p\mathbf{g}(\hat{\mathbf{x}})/dt^p$ of the visible states are generated from the symbolic model and compared with finite difference derivatives $\Delta^p\mathbf{g}(\mathbf{x})/\Delta t^p$ computed directly from the sequence of visible states. The entire architecture is trained end-to-end using the mean squared error (MSE) loss between the symbolic and finite difference derivatives.}
\end{figure*}

We propose a machine learning framework for solving the common problem of partially observed system identification, where a portion of the system state is observed but the remaining hidden states as well as the underlying dynamics are unknown. Unlike in the generic high dimensional setting, this is a much more structured problem, and we take full advantage of this additional structure when designing our architecture. To deal with having only partial state information, our method combines an encoder, for reconstructing the full system state, and a sparse symbolic model, which learns the system dynamics, providing a flexible framework for both system identification and state reconstruction (Fig.\ \ref{fig:arch}). The full architecture is trained by matching the higher order time derivatives of the symbolic model with finite difference estimates from the data. As illustrated in our numerical experiments, this approach can be easily adapted for specific applications by incorporating known constraints into the architecture of the encoder and the design of the symbolic model.

\emph{Problem Formulation.}---Consider a nonlinear dynamical system defined by the first order ODE
\begin{equation}
    \frac{d\mathbf{x}}{dt} = \mathbf{F}(\mathbf{x}).
\end{equation}
The visible or observed state is given by a known ``projection'' function $\mathbf{x}_v = \mathbf{g}(\mathbf{x})$ while the hidden states $\mathbf{x}_h$ must be reconstructed such that $\mathbf{a}(\mathbf{x}_v,\mathbf{x}_h) = \mathbf{x}$, where $\mathbf{a}$ is a known aggregation function. The goal is to determine the governing equations defined by $\mathbf{F}(\mathbf{x})$ while simultaneously reconstructing the hidden state $\mathbf{x}_h$.

Without prior knowledge detailing the structure of the dynamical system, we can generically choose the visible state $\mathbf{x}_v = (x_1, x_2,\ldots,x_k)$ to be a subset of the full state $\mathbf{x} = (x_1, x_2,\ldots,x_k,x_{k+1},\ldots,x_n)$, i.e.\ $\mathbf{g}$ is a simple projection of $\mathbf{x}$ onto the subset $\mathbf{x}_v$. The remaining components would then form the hidden state $\mathbf{x}_h = (x_{k+1},x_{k+2},\ldots,x_n)$, and the aggregation function $\mathbf{a}$ just concatenates of the two states $\mathbf{x}_v,\mathbf{x}_h$. When additional information about the dynamical system is available, $\mathbf{g}$ and $\mathbf{a}$ can be chosen appropriately to reflect the structure of the dynamics (e.g.\ see our nonlinear Schr\"{o}dinger phase reconstruction example).

\emph{Proposed Machine Learning Framework.}---Our proposed framework consists of an encoder, which uses the visible states to reconstruct the corresponding hidden states, and an interpretable symbolic model, which represents the governing equations of the dynamical system. The encoder $\mathbf{e}_\eta$, typically a neural network architecture with learnable parameters $\eta$, takes as input the sequence of visible states $\{\mathbf{x}_v(t_0),\mathbf{x}_v(t_0 + \Delta t),\ldots,\mathbf{x}_v(t_N)\}$ and reconstructs the hidden states $\{\hat{\mathbf{x}}_h(t_0),\hat{\mathbf{x}}_h(t_0 + \Delta t),\ldots,\hat{\mathbf{x}}_h(t_N)\}$. We can then obtain a reconstruction of the full state by applying the aggregation function $\hat{\mathbf{x}} = \mathbf{a}(\mathbf{x}_v,\hat{\mathbf{x}}_h)$. The fully reconstructed state $\hat{\mathbf{x}}$ allows us to compute symbolic time derivatives defined by a symbolic model of the governing equations
\begin{equation}
    \frac{d\hat{\mathbf{x}}}{dt} = \hat{\mathbf{F}}_\theta(\hat{\mathbf{x}}) \coloneqq \theta_1 \mathbf{f}_1(\hat{\mathbf{x}}) + \theta_2 \mathbf{f}_2(\hat{\mathbf{x}}) + \cdots + \theta_m \mathbf{f}_m(\hat{\mathbf{x}}),
\label{eq:sym}
\end{equation}
where $\theta_1,\theta_2,\ldots,\theta_m$ are learnable coefficients and $\mathbf{f}_1,\mathbf{f}_2,\ldots,\mathbf{f}_m$ are predefined terms, such as monomial expressions or linear combinations representing spatial derivatives (for PDE systems). The dimensionality of the system state $\mathbf{x}$ is also a hyperparameter.

To jointly train the encoder and symbolic model using only partial observations, we match higher order time derivatives of the visible states with finite difference estimates from the data. These time derivatives are implicitly defined by the symbolic model (Eq.\ \ref{eq:sym}), so we develop and use an algorithmic trick that allows standard automatic differentiation methods \cite{JMLR:v18:17-468} to compute higher order symbolic time derivatives of the {\em reconstructed} visible states $\mathbf{g}(\hat{\mathbf{x}})$ (see Supplemental Materials). These symbolic derivatives can then be compared with finite difference time derivatives $\Delta^p \mathbf{g}(\mathbf{x})/\Delta t^p = \Delta^p\mathbf{x}_v/\Delta t^p$ computed directly from the visible states $\mathbf{x}_v$.

We train the entire architecture in an end-to-end fashion by optimizing the mean squared error (MSE) loss
\begin{equation}
    \mathcal L(\eta,\theta) = \frac{1}{N}\sum_{i=1}^N\sum_{p=1}^{M}\alpha_p\left(\frac{d^p\mathbf{g}(\hat{\mathbf{x}}(t_i))}{dt^p} - \frac{\Delta^p\mathbf{x}_v(t_i)}{\Delta t^p}\right)^2,
\end{equation}
where $\alpha_p$ are hyperparameters that determine the importance of each derivative order in the loss function. This loss implicitly depends on the encoder $\mathbf{e}_\eta$ through the reconstructed state $\hat{\mathbf{x}}$ and the symbolic model $\hat{\mathbf{F}}_\theta$ through the symbolic time derivatives. To achieve sparsity in the symbolic model, we use a simple thresholding approach---commonly used in sparse linear regression applications \cite{Brunton3932}---which sets a coefficient $\theta_i$ to zero if its absolute value falls below a chosen threshold $\theta_\mathrm{thres}$. We implement this sparsification at regular intervals during training. See the Supplemental Materials for additional architecture and training details.

The code for implementing our framework and reproducing our results is available at \url{https://github.com/peterparity/symder}.

\begin{figure*}
\includegraphics{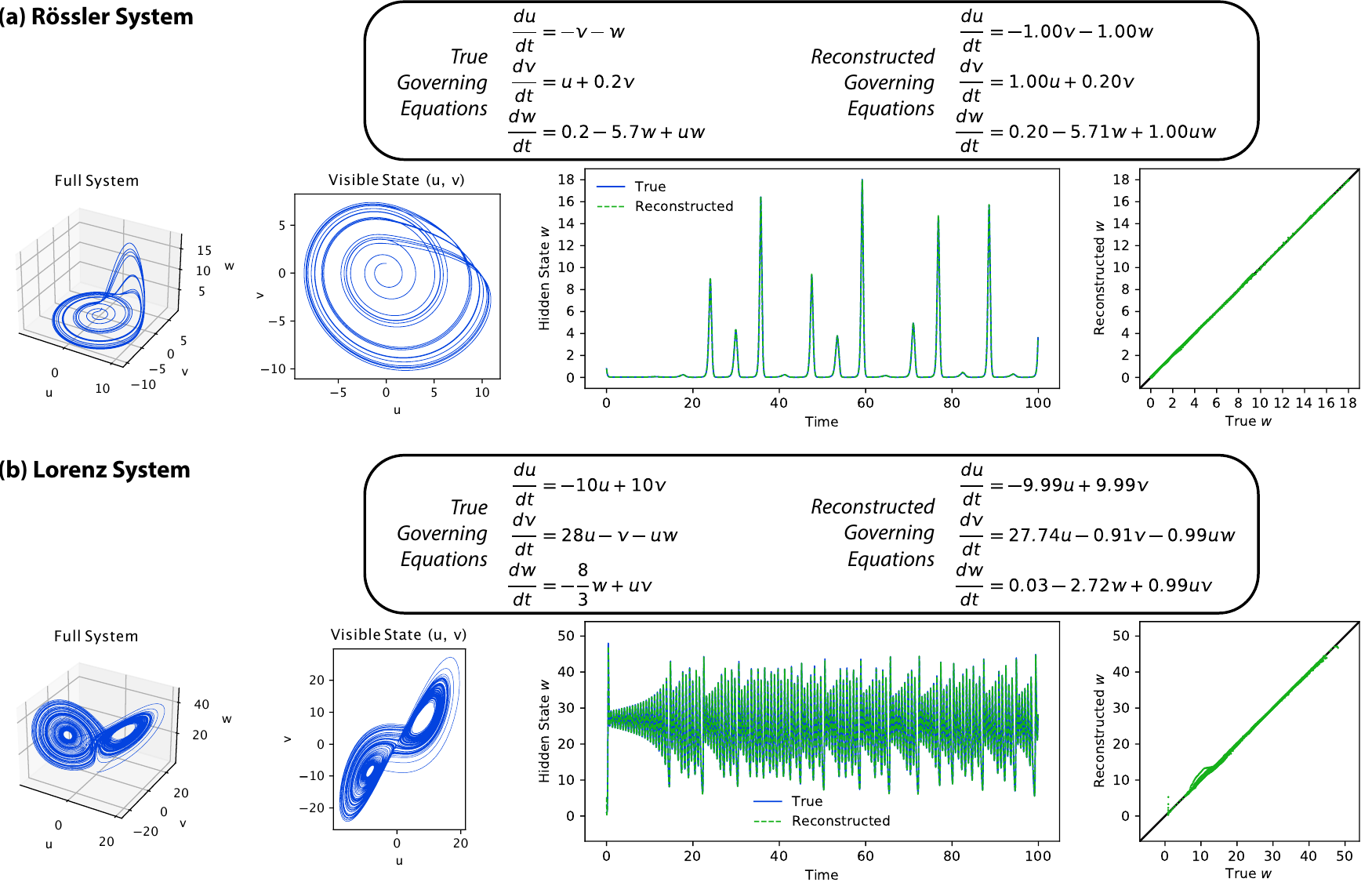}
\caption{\label{fig:ode}System identification and hidden state reconstruction for the (a) R\"{o}ssler and (b) Lorenz systems. In both numerical experiments, the $u$ and $v$ components are visible while the $w$ component is hidden. The true and reconstructed hidden states $w$ are shown as a function of time and also plotted directly against each other for comparison.}
\end{figure*}

\begin{figure*}
\includegraphics{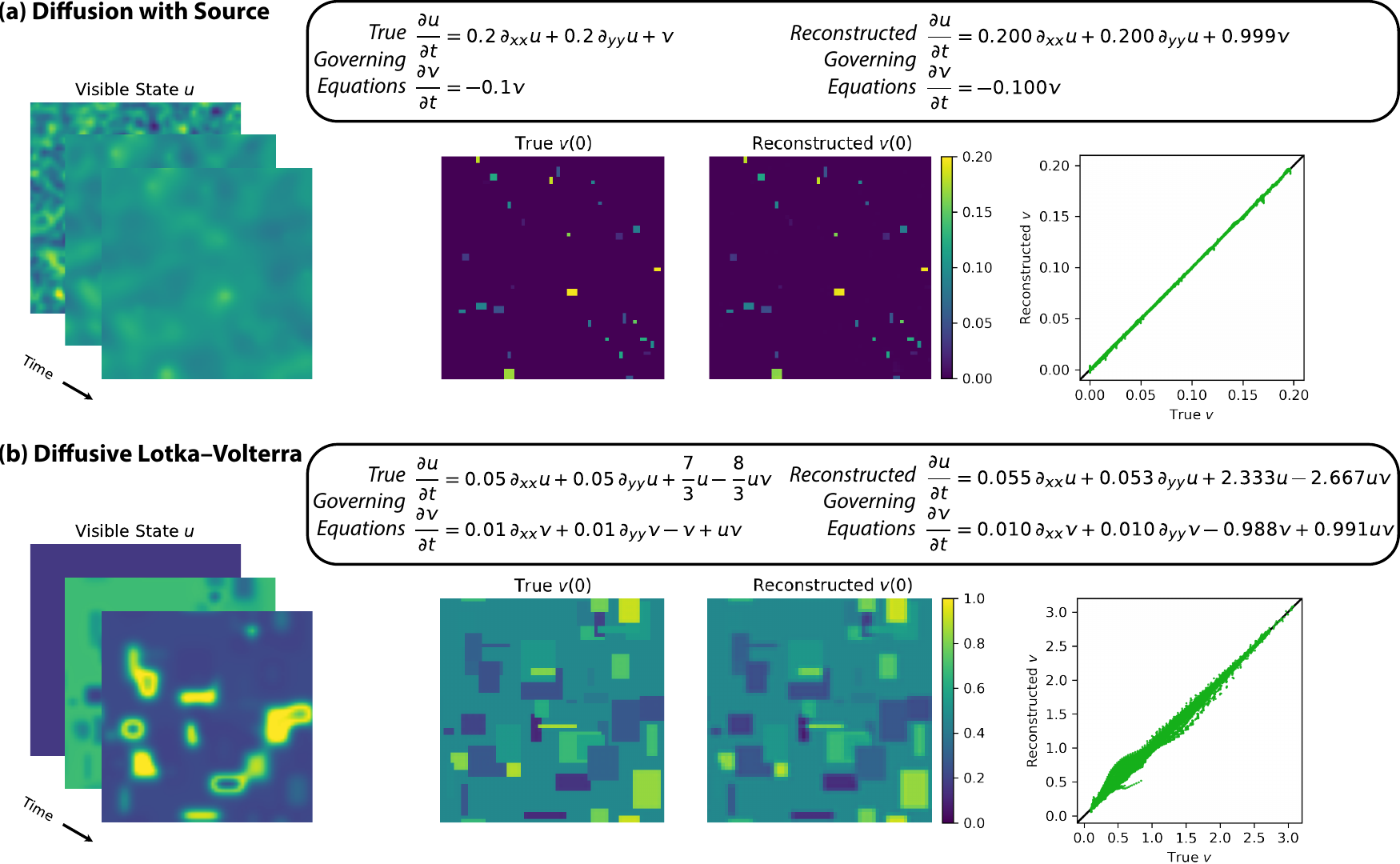}
\caption{\label{fig:pde}System identification and hidden state reconstruction for the (a) diffusion system with a decaying source term $v$ and (b) diffusive Lokta--Volterra system. In both numerical experiments, the $u$ component is visible while the $v$ component is hidden. The true and reconstructed hidden states $v$ are shown at time $t=0$ and are also plotted directly against each other for comparison.}
\end{figure*}

\emph{ODE Experiments.}---To demonstrate our method, we use data from two standard examples of chaotic nonlinear dynamics: the R\"{o}ssler system (Fig.\ \ref{fig:ode}a) and the Lorenz system (Fig.\ \ref{fig:ode}b). Both systems have a three-dimensional phase space $(u,v,w)$, and we take the first two dimensions $(u,v)$ to be the visible state with the remaining dimension $w$ as the hidden state. In both cases, we are able to accurately identify the governing equations and reconstruct the hidden state $w$ (Fig.\ \ref{fig:ode}). We achieve a relative reconstruction error of $4.6\times 10^{-4}$ (relative to the range of the hidden state) for the R\"{o}ssler system and $1.7\times 10^{-3}$ for the Lorenz system.

\emph{PDE Experiments.}---To test our method in a more challenging setting, we use data from two PDE systems: a 2D diffusion system with an exponentially decaying source term (Fig.\ \ref{fig:pde}a) and a 2D diffusive Lokta--Volterra predator--prey system (Fig.\ \ref{fig:pde}b)---commonly used for ecological modeling \cite{DUBOIS197567,COMINS197475,KMET1994277}. For the diffusion system, we observe a diffusing visible state $u(x,y,t)$ and must reconstruct the hidden dynamic source term $v(x,y,t)$. Similarly, for the diffusive Lokta--Volterra system, one of the two components is visible $u(x,y,t)$ while the other is hidden $v(x,y,t)$. We accurately identify the governing equations and reconstruct the hidden component for both systems (Fig.\ \ref{fig:pde}), achieving a relative error of $1.4\times 10^{-4}$ for the diffusion system and $1.0\times 10^{-3}$ for the diffusive Lokta--Volterra system. The neural network encoder has more difficulty with the more complex and nonlinear diffusive Lokta--Volterra system, resulting in a slightly blurry reconstruction.

\begin{figure}
\includegraphics{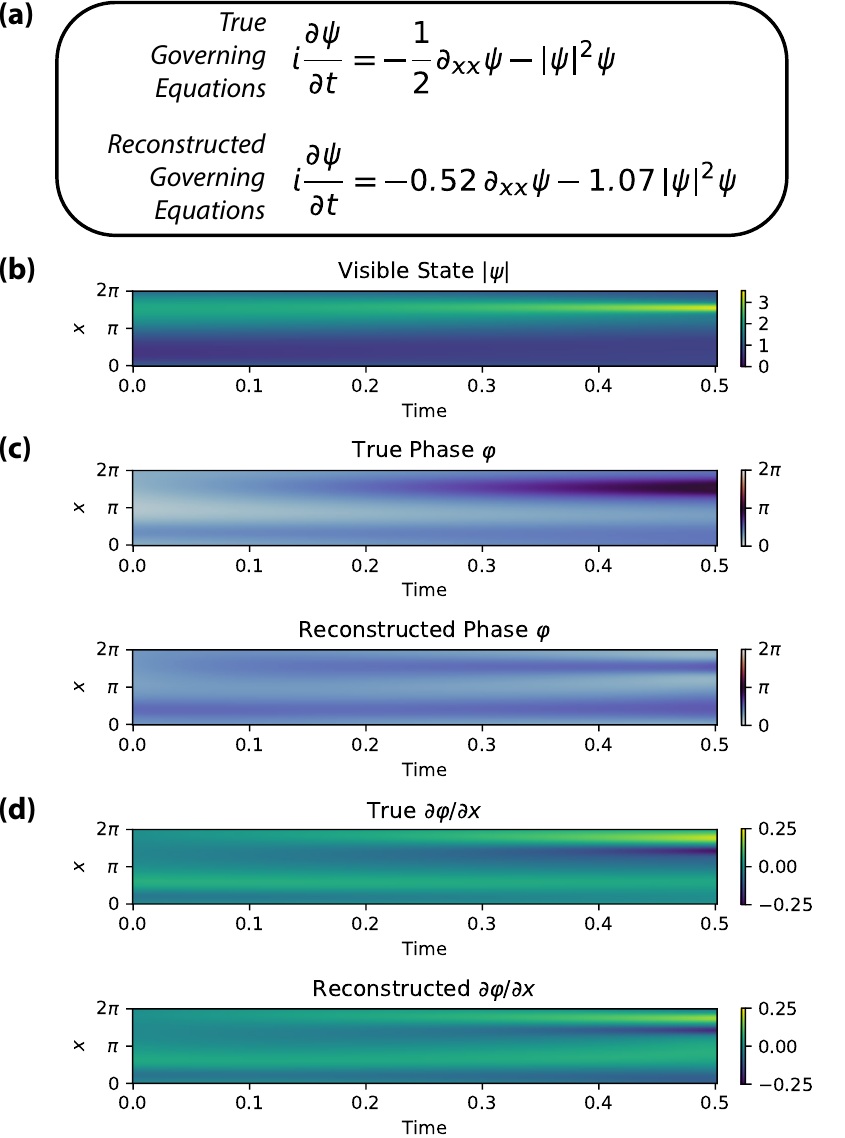}
\caption{\label{fig:nlse}System identification (a) and phase reconstruction (c,d) for the nonlinear Schr\"{o}dinger system. The magnitude $|\psi|$ of the wave is visible (b) while the phase $\varphi = \mathrm{arg}(\psi)$ is hidden (c) and must be reconstructed. The spatial derivative of the phase $\partial\varphi/\partial x$ (d) and its reconstruction are also shown.}
\end{figure}

\emph{Phase Reconstruction.}---As a final example, we consider the phase reconstruction problem for the 1D nonlinear Schr\"{o}dinger equation---a model for light propagation through a nonlinear fiber \cite{ablowitz_2011}---to demonstrate the breadth of our approach and its ability to handle a more difficult and structured problem. Using only visible amplitude data $|\psi(x,t)|$, we aim to identify the underlying dynamics and reconstruct the hidden phase $\varphi(x,t) = \mathrm{arg}(\psi(x,t))$. For this system, we also assume we have some prior knowledge about the structure of the dynamics: a complex wave equation with a global phase shift symmetry and only odd nonlinearities to model an optical material with inversion symmetry \cite{ablowitz_2011}. This allows us to limit the library of predefined terms used by our symbolic model. Our prior knowledge also informs our choice of projection $\mathbf{g}(\psi) = |\psi|$ and aggregation functions $\mathbf{a}(|\psi|, \varphi) = |\psi|e^{i\varphi}$.

Our method successfully identifies the governing equation for the nonlinear Schr\"{o}dinger data and roughly captures the correct phase profile. Although the overall phase reconstruction seems somewhat poor, with a relative error of 0.35, this also includes an accumulated drift of the phase over time. The spatial derivative of the phase $\partial\varphi/\partial x$ has a much more reasonable relative error of 0.057. Furthermore, given the governing equations extracted by our method, other more specialized algorithms for nonlinear phase retrieval can be used as a post-processing step to significantly improve the quality of the phase reconstruction \cite{Lu:13}.

\emph{Conclusion.}---On a wide variety of dynamical systems, we have demonstrated that our proposed machine learning framework can successfully identify sparse interpretable dynamics and reconstruct hidden states using only partial observations. By fitting symbolic models, we are able to discover the exact form of the symbolic equations governing the underlying physical systems, resulting in highly interpretable models and predictions (see Supplemental Materials). Our method is also straightforward to implement and use, easily adapting to differing levels of prior knowledge about the unknown hidden states and dynamics.

Compared with methods that require explicit integration \cite{NEURIPS2018_69386f6b,9053035}, our approach can be significantly more computational efficient since we only need to compute symbolic and finite difference derivatives. Methods that rely on explicit integration may also need to deal with stiffness and other issues that are relevant to choosing an appropriate integration scheme \cite{2001.04385}. However, methods using explicit integration also have the advantage of being much more robust to noise. Because we require higher order finite difference time derivative estimates from data, our approach---like other derivative-based methods---is generally more susceptible to noise. Careful tuning of our sparsity method helps mitigate this to some extent in a similar fashion to methods like SINDy \cite{Brunton3932,doi:10.1098/rspa.2020.0279}, and promising new methods for identifying the noise distribution alongside the dynamics \cite{2009.08810} could be incorporated into our framework in the future.

Our framework offers a strong foundation for designing interpretable machine learning methods to deal with partial observations and solve the combined system identification and state reconstruction task. We hope to continue developing more robust encoders and more flexible symbolic models that will work within our proposed framework. For example, the encoder (see Supplemental Materials) used in our final experiment on phase reconstruction has similarities with variational approaches used for PDE discovery \cite{RAISSI2019686,2005.03448}, and we believe that these variational methods can be incorporated into our framework to provide a smoother encoding and improve robustness to noise. In future work, we will also study symbolic models that have multiple layers of composable units designed for symbolic regression tasks \cite{9180100,2007.10784,Udrescueaay2631}. These alternative symbolic architectures provide more powerful and flexible models with a sparse symbolic prior, potentially addressing some current limitations of our implementation (see Supplemental Materials) and allowing our framework to handle a wider range of governing equations---such as the Hill equations used in modeling gene expression \cite{alon2019introduction}---without requiring large libraries of predefined terms.

\begin{acknowledgments}
We would like to acknowledge useful discussions with Samuel Kim, Rumen Dangovski, Charlotte Loh, Andrew Ma, and Ileana Rugina. This research is supported in part by the U.S.\ Department of Defense through the National Defense Science \& Engineering Graduate Fellowship (NDSEG) Program. This work is further supported in part by the National Science Foundation under Cooperative Agreement PHY-2019786 (The NSF AI Institute for Artificial Intelligence and Fundamental Interactions, \url{http://iaifi.org/}). It is also based upon work supported in part by the U.S.\ Army Research Office through the Institute for Soldier Nanotechnologies at MIT, under Collaborative Agreement Number W911NF-18-2-0048. Research was also sponsored in part by the United States Air Force Research Laboratory and the United States Air Force Artificial Intelligence Accelerator and was accomplished under Cooperative Agreement Number FA8750-19-2-1000. The views and conclusions contained in this document are those of the authors and should not be interpreted as representing the official policies, either expressed or implied, of the United States Air Force or the U.S.\ Government. The U.S.\ Government is authorized to reproduce and distribute reprints for Government purposes notwithstanding any copyright notation herein.
\end{acknowledgments}



\bibliography{biblio}

\pagebreak
\onecolumngrid
\begin{center}
\textbf{\large Supplemental Materials: Discovering Sparse Interpretable Dynamics from Partial Observations}
\end{center}
\setcounter{equation}{0}
\setcounter{figure}{0}
\setcounter{table}{0}
\setcounter{page}{1}
\makeatletter
\renewcommand{\theequation}{S\arabic{equation}}
\renewcommand{\thefigure}{S\arabic{figure}}
\twocolumngrid

\section{Automatic Computation of Symbolic Derivatives}
The time derivatives can be derived by repeated differentiation of the symbolic model
\begin{equation}
    \frac{d\hat{\mathbf{x}}}{dt} = \hat{\mathbf{F}}_\theta(\hat{\mathbf{x}}),
\label{eq:sym1}
\end{equation}
substituting back in previously computed derivatives to obtain expressions only in terms of the reconstructed state $\hat{\mathbf{x}}$. For example, the first and second time derivatives can be written in index notation as
\begin{align}
    \frac{dg_i}{dt} &= \sum_j\frac{dg_i}{d\hat{x}_j}\frac{d\hat{x}_j}{dt} = \sum_j\frac{dg_i}{d\hat{x}_j}\hat{F}_{\theta j} \label{eq:deriv1}\\
    \frac{d^2g_i}{dt^2} &= \sum_{j,k}\frac{d^2g_i}{d\hat{x}_jd\hat{x}_k}\hat{F}_{\theta j}\hat{F}_{\theta k} + \frac{dg_i}{d\hat{x}_j}\frac{d\hat{F}_{\theta j}}{d\hat{x}_k}\hat{F}_{\theta k}. \label{eq:deriv2}
\end{align}

The expressions for the symbolic time derivatives (Eqs.\ \ref{eq:deriv1} \& \ref{eq:deriv2}) quickly grow more and more unwieldy for higher order derivatives. Implementing these expressions by hand is likely to be both time-consuming and error-prone, especially for more complex symbolic models such as those used in our PDE experiments. To address this issue, we develop an automated approach that takes advantage of powerful modern automatic differentiation software (in our case, the JAX library \cite{jax2018github}).

Automatic differentiation is the algorithmic backbone of modern deep learning \cite{JMLR:v18:17-468}, and a new generation of source-to-source automatic differentiation libraries are quickly becoming available \cite{jax2018github,1810.07951}. Automatic differentiation uses a library of custom derivative rules defined on a set of primitive functions which can then be arbitrarily composed to form more complex expressions. The algorithm normally requires a forward evaluation of a function that sets up a backward pass which computes the gradient of the function. In our case, the appropriate forward step is integrating the symbolic model (Eq.\ \ref{eq:sym1}) using an ODE solver, which makes the time variable and its derivatives explicit rather than being implicitly defined by the governing equations. This, however, would introduce significant overhead and would not produce the exact expressions that we derived earlier. In fact, integration should not be necessary at all for efficiently implementing symbolic differentiation. Instead, we propose a simple algorithmic trick that allows standard automatic differentiation to compute symbolic time derivatives without explicit integration.

Consider a function $\mathcal I(\hat{\mathbf{x}}, \epsilon)$ that propagates the state $\hat{\mathbf{x}}$ forward by a time $\epsilon$ according to the governing equations (Eq.\ \ref{eq:sym1}), i.e.\ 
\begin{equation}
    \mathcal I(\hat{\mathbf{x}}(t), \epsilon) = \mathbf{\hat{\mathbf{x}}}(t+\epsilon).
\end{equation}
As $\epsilon \to 0$, $\mathcal I(\hat{\mathbf{x}}(t), 0) = \hat{\mathbf{x}}(t)$ reduces to the identity. Taking a derivative with respect to $\epsilon$, we find that
\begin{align}
    \begin{split}
    \frac{\partial \mathcal I(\hat{\mathbf{x}}(t), \epsilon)}{\partial \epsilon} &= \frac{d\mathbf{\hat{\mathbf{x}}}(t+\epsilon)}{d\epsilon} \\
    &= \hat{\mathbf{F}}_\theta(\hat{\mathbf{x}}(t+\epsilon)) \\
    &= \hat{\mathbf{F}}_\theta(\mathcal I(\hat{\mathbf{x}}(t), \epsilon)),
    \end{split}
\end{align}
which reduces to $\partial\mathcal I(\hat{\mathbf{x}},\epsilon)/\partial\epsilon|_{\epsilon = 0} = d\hat{\mathbf{x}}/dt = \hat{\mathbf{F}}_\theta(\hat{\mathbf{x}})$ as $\epsilon \to 0$. This generalizes to higher order derivatives, allowing us to compute time derivatives of $\hat{\mathbf{x}}$ as
\begin{equation}
    \frac{d^p\hat{\mathbf{x}}}{dt^p} = \left.\frac{\partial^p\mathcal I(\hat{\mathbf{x}}, \epsilon)}{\partial\epsilon^p}\right|_{\epsilon=0}.
\end{equation}
Since we only ever evaluate at $\epsilon = 0$, this formulation makes the time variable explicit without having to integrate the governing equations. To implement this trick using an automatic differentiation algorithm, we define a wrapper function $\mathcal I_0(\hat{\mathbf{x}}, \epsilon) \coloneqq \hat{\mathbf{x}}$ that acts as the identity on the state $\hat{\mathbf{x}}$ but has a custom derivative rule
\begin{equation}
    \frac{\partial \mathcal I_0(\hat{\mathbf{x}}, \epsilon)}{\partial\epsilon} \coloneqq \hat{\mathbf{F}}_\theta(\mathcal I_0(\hat{\mathbf{x}}, \epsilon)).
\end{equation}
This allows standard automatic differentiation to correctly compute exact symbolic time derivatives of our governing equations, including higher order derivatives. Our code for implementing this algorithmic trick and for reproducing the rest of our results is available at \url{https://github.com/peterparity/symder}.

The proposed algorithmic trick for computing higher order time derivatives, which exploits modern automatic differentiation, further simplifies the implementation of our method and allows the user to focus on designing an appropriate encoder and choosing a reasonable library of predefined terms for the sparse symbolic model.

\section{Dataset, Architecture, and Training Details}
The data and architecture requirements for using our approach are dependent on the properties of the dynamical system, the fraction of visible states, and the chosen symbolic model. For example, a more constrained symbolic model with a smaller library of terms will likely be more data efficient due to having a stronger inductive bias on the model. In general, the trajectories from the data need to be long and varied enough in order to differentiate among the terms provided by the symbolic model, although we have found that a single trajectory is often sufficient for accurate state reconstruction and system identification. For the encoder architecture, we generally use small and relatively shallow neural networks, which already provide good hidden state reconstruction performance. This also means that our approach trains reasonably quickly, taking $\sim$ 2.5 minutes for the ODE systems on a single consumer GPU (GeForce RTX 2080 Ti) and $\sim$ 2 hours for the much larger PDE systems on four GPUs. However, it is certainly possible that more structured encoders may help in certain cases requiring more complex reconstructions. In addition, we are able to obtain accurate results in our tests by matching the first and second order time derivatives, although higher order derivatives will be necessary for datasets with a larger fraction of hidden states.

\subsection{ODE Systems}
Each system is sampled for 10000 time steps of size $\Delta t = 10^{-2}$, and the resulting time series data and computed finite difference derivatives are normalized to unit variance.

The encoder takes a set of nine visible states $\{\mathbf{x}_v(t-4\Delta t),\mathbf{x}_v(t-3\Delta t),\ldots,\mathbf{x}_v(t+4\Delta t)\}$ as input to reconstruct each hidden state $\hat{\mathbf{x}}_h(t)$ and is implemented as a sequence of three 1D time-wise convolutional layers with kernel sizes 9--1--1 and layer sizes 128--128--1. This architecture enforces locality in time, allowing the neural network to learn a simpler and more interpretable mapping. The predefined terms of the symbolic model consist of constant, linear, and quadratic monomial terms, i.e.\ $1$, $u$, $v$, $w$, $u^2$, $v^2$, $w^2$, $uv$, $uw$, and $vw$, for each governing equation.

We also scale the effective time step of the symbolic model by a factor of $10$ to improve training by preconditioning the model coefficients. We then train for 50000 steps using the AdaBelief optimizer \cite{NEURIPS2020_d9d4f495} with learning rate $10^{-3}$ and with hyperparameters $\alpha_1 = \alpha_2 = 1$ to equally weight the first two time derivative terms in the loss function ($\alpha_p = 0$ for $p > 2$). Every 5000 training steps, we sparsify the symbolic model, setting coefficients to zero if their absolute value is below $\theta_\mathrm{thres} = 10^{-3}$.

One additional caveat is that the equation and hidden state obtained by our approach is not exactly the same as the original and instead corresponds to the correct governing equations for an affine transformation of the hidden state $w' = aw + b$. In order to make a direct comparison, we use linear regression to fit the reconstructed hidden states to the original hidden states and show the resulting transformed equations.

\subsection{PDE Systems}
Each system is sampled on a $64\times64$ spatial mesh with grid spacing $\Delta x = \Delta y = 1$ for 1000 time steps of size $\Delta t = 5\times10^{-2}$, and the resulting data and estimated derivatives are normalized to unit variance.

The encoder is a sequence of three 3D spatiotemporal convolutional layers with kernel sizes 5--1--1 and layer sizes 64--64--1, which enforces locality in both time and space. The predefined terms of the symbolic model consist of constant, linear, and quadratic terms as well as up to second order spatial derivative terms, e.g.\ $\partial_x u$, $\partial_y u$, $\partial_{xx} u$, $\partial_{yy} u$, $\partial_{xy} u$, and similarly for $v$.

We scale the effective time step and spatial grid spacing of the symbolic model by a factor of $10$ and $\sqrt{10}$, respectively, to precondition the model coefficients. For the diffusion system, we train for 50000 steps with learning rate $10^{-4}$ and hyperparameters $\alpha_1 = 1$ and $\alpha_2 = 10$, and we sparsify the symbolic model every 1000 training steps with $\theta_\mathrm{thres} = 5\times 10^{-3}$. For the diffusive Lokta--Volterra system, we train for 100000 steps with learning rate $10^{-3}$ and hyperparameters $\alpha_1 = \alpha_2 = 1$, and we sparsify the symbolic model every 1000 training steps with $\theta_\mathrm{thres} = 2\times 10^{-3}$.

\subsection{Phase Reconstruction}
The system is sampled on a size 64 mesh with spacing $\Delta x = 2\pi/64$ for 500 time steps of size $\Delta t = 10^{-3}$.

Using the available prior knowledge, we allow the symbolic model to use spatial derivative terms $\partial_x^p \psi$ for $p\in\{1,2,3,4\}$ and nonlinearity terms $|\psi|^q\psi$ for $q=\{2,4,6,8\}$. We scale the effective time step by a factor of $10$ to precondition the model coefficients and train for 100000 time steps with learning rate $10^{-4}$ and hyperparameters $\alpha_1 = \alpha_2 = 1$ and $\beta = 10^3$. We sparsify the symbolic model every 10000 training steps with $\theta_\mathrm{thres} = 10^{-3}$.

Unlike the previous examples, reconstructing the phase is a much trickier problem that cannot be done using a local spatiotemporal encoder. Instead of using a neural network mapping, we use a direct embedding of the phase as function of time, i.e.\ for each point in the spatiotemporal grid of the original data, we learn a parameter for the phase $\hat\varphi(x,t)$. This simple approach has the advantage of being incredibly flexible but also more difficult to train, requiring an additional encoder regularization term
\begin{equation}
    \mathcal R_\mathrm{enc} = \beta\left(\frac{\partial\hat\psi}{\partial t} - \frac{\Delta \hat\psi}{\Delta t}\right)^2
\end{equation}
that ensures the symbolic time derivatives match the finite difference time derivatives of the reconstructed state $\hat\psi = \mathbf{a}(|\psi|, \hat\varphi)$. Unlike the compact neural network encoders from the previous experiments, this encoder also must scale with the dataset size and does not provide a useful mapping that can be used for future hidden state reconstruction.

\begin{figure}
\includegraphics[width=3.38in]{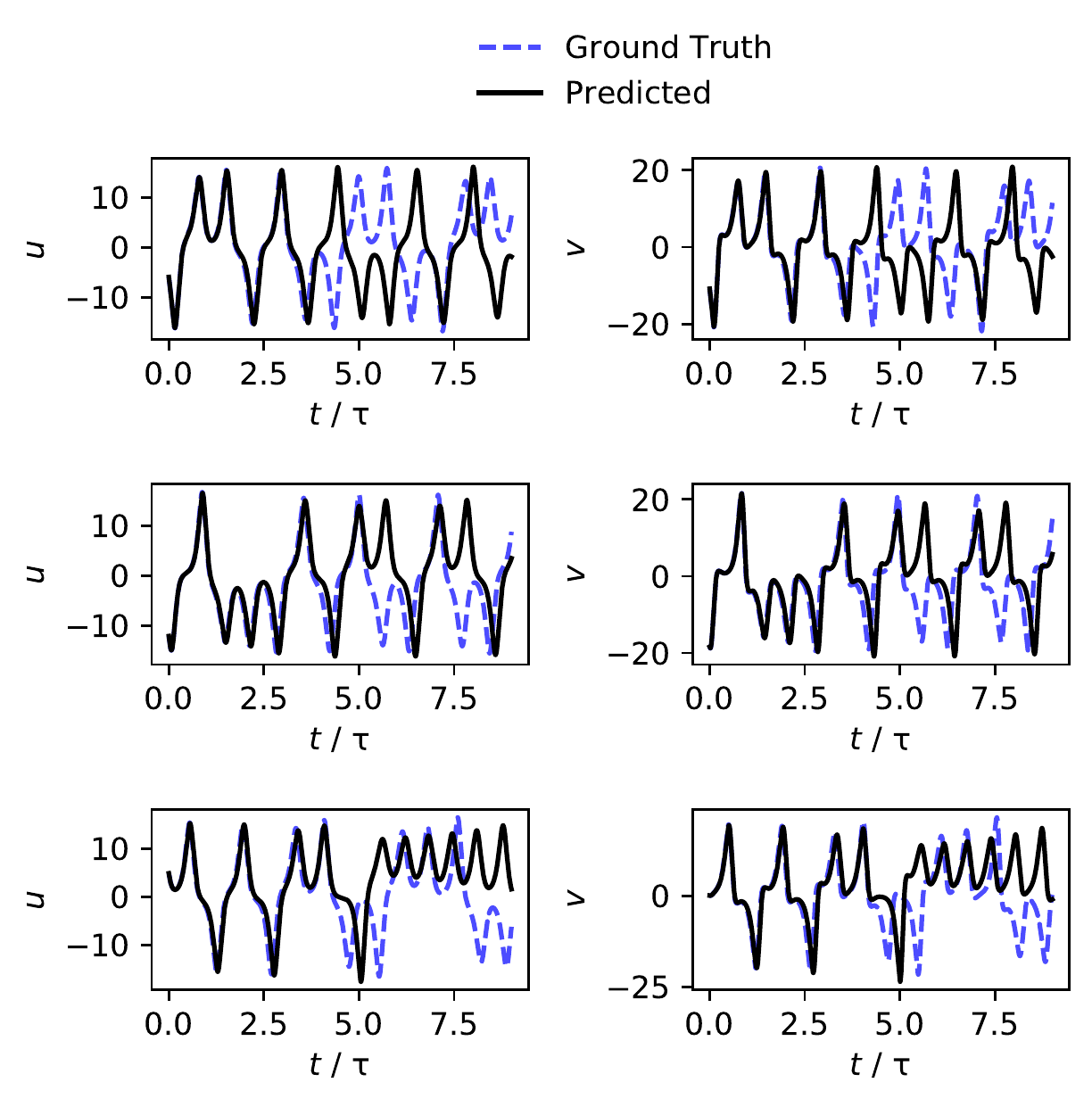}
\caption{\label{fig:lorenz_pred}Three examples of prediction on test trajectories from the model trained on the Lorenz system with two visible states $u$, $v$. The time is given in units of the Lyapunov time $\tau$, the characteristic time for two nearby trajectories to exponentially diverge in a chaotic system.}
\end{figure}

\begin{figure}
\includegraphics{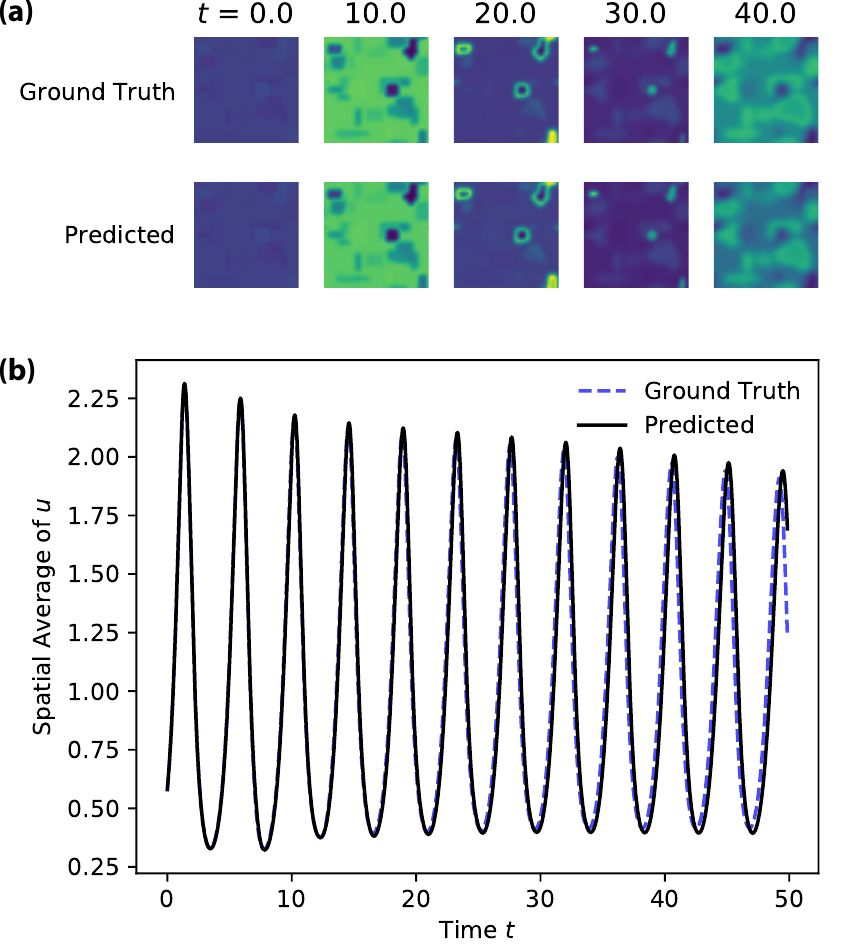}
\caption{\label{fig:reacdiff_pred} (a) Snapshots of the predicted vs.\ ground truth visible state $u(x,y,t)$ from the model trained on the diffusive Lokta--Volterra system with a single visible state. (b) Predicted vs.\ ground truth spatial average of the visible state $u$ as a function of time.}
\end{figure}

\section{Prediction Examples}
Because we are able to capture the true dynamics of each system using a symbolic model, our models exhibit excellent generalization performance beyond the training data. The expected accuracy of the predictions from the model is also highly interpretable, with prediction performance depending on the accuracy of the fitted coefficients of the symbolic model and the hidden state reconstruction of the initial state. As a result, it is often more instructive to directly examine the coefficients of the symbolic model and the quality of the hidden state reconstruction when judging the success of our method, both of which are discussed in the main text. Here, for illustrative purposes, we show prediction examples from the Lorenz system (Fig.\ \ref{fig:lorenz_pred}) and the diffusive Lokta--Volterra system (Fig.\ \ref{fig:reacdiff_pred}), demonstrating the quality and interpretability of the prediction performance that is expected from well-trained symbolic models.

Because the Lorenz system is chaotic, long-term prediction performance has a theoretically limit characterized by the Lyapunov time $\tau$ of the system. Despite this, we are still able to predict well up to $\sim 4\tau$ and capture the correct behavior of the system using our symbolic model (Fig.\ \ref{fig:lorenz_pred}). Prediction performance on the diffusive Lokta--Volterra system is also very good (Fig.\ \ref{fig:reacdiff_pred}) with the slight deviations at long times resulting from imperfections in the initial hidden state reconstruction and small errors in the learned coefficients of the symbolic model. These are highly interpretable sources of error that are easy to analyze and iteratively refine.

\section{Limitations of Reconstructing Hidden States}
While we have demonstrated that our approach performs well over a wide range of tasks, there are still limitations to its ability to reconstruction hidden states and therefore identify the correct symbolic models. General theoretical limitations include the degree of interaction between the visible and hidden states as well as measurement noise. If a hidden state does not significantly affect the dynamics of the available visible states, it will be very difficult or impossible for any method to reconstruct. Furthermore, for data with a smaller fraction of visible states and thus more hidden states, the reconstruction task becomes harder. This manifests itself in our framework as requiring higher and higher order time derivative matching in order to fully reconstruct the hidden states, resulting in more sensitivity to noise. For example, if we want to use no more than second order time derivatives in our loss function, we require at least $N/2$ visible states for an $N$-dimensional system.

There are also more subtle issues with sparse optimization that causes problems for our current implementation. For example, if we attempt to reconstruct the Rossler or Lorenz system using only a single visible state, our symbolic model tends to get stuck in local minima and fails to find the correct sparsity pattern, i.e.\ the right sparse combination of terms. However, if we provide the correct sparsity pattern to the symbolic model, we obtain very accurate results on par with what we have shown for the case of two visible states. That is, given the right sparsity pattern, our method is able to learn an accurate reconstruction of the hidden states and also fit the coefficients of the symbolic model using only a single visible state. On the one hand, this highlights the advantages of having strong inductive biases, which can help avoid some of these bad local minima. On the other hand, there is a clear need for a more sophisticated approach to sparsity and symbolic regression \cite{9180100,2007.10784,Udrescueaay2631}, which we hope to explore further in the future.

\end{document}